\documentclass{article}

\usepackage{arxiv}

\usepackage[T1]{fontenc}    
\usepackage{hyperref}       
\usepackage{url}            
\usepackage{booktabs}       
\usepackage{amsfonts}       
\usepackage{nicefrac}       
\usepackage{microtype}      
\usepackage{lipsum}
\usepackage{graphicx} 
\graphicspath{{./images/}}
\title{LimeOut: An Ensemble Approach To Improve Process Fairness}

\author{
  Vaishnavi Bhargava \\
Universit\'e de Lorraine,\\
 CNRS, Inria, LORIA, F-54000 Nancy, France\\
  \texttt{vaishnavi.bhargava2605@gmail.com} \\
   \And
 Miguel Couceiro \\
  Universit\'e de Lorraine,\\
  CNRS, Inria, LORIA, F-54000 Nancy, France\\
  \texttt{miguel.couceiro@loria.fr} \\
   \And
  Amedeo Napoli \\
  Universit\'e de Lorraine,\\
  CNRS, Inria, LORIA, F-54000 Nancy, France\\
  \texttt{amedeo.napoli@loria.fr} 
}

\begin{document}
\maketitle
\begin{abstract}
Artificial Intelligence and Machine Learning are becoming increasingly present in several aspects of human life, especially, those dealing with decision making. Many of these algorithmic decisions are taken without human supervision and through decision making processes that are not transparent.
This raises concerns regarding the potential bias of these processes towards certain groups of society, which may entail unfair results and, possibly, violations of human rights. Dealing with such biased models is one of the major concerns to maintain the public trust. 

In this paper, we address the question of {\it process} or {\it procedural fairness}. More precisely, we consider the problem of making classifiers fairer by reducing their dependence on sensitive features while increasing (or, at least, maintaining) their accuracy. To achieve both, we  draw inspiration from ``dropout'' techniques in neural based approaches, and propose a framework that relies on ``feature drop-out'' to tackle process fairness. We make use of ``LIME Explanations'' to assess a classifier's fairness and to determine the sensitive features to remove. This produces a pool of classifiers (through feature dropout) whose ensemble is shown empirically to be less dependent on sensitive features, and  with improved or no impact on accuracy. 
\end{abstract}

\keywords{Explainability \and Fairness \and Feature importance \and Feature-dropout \and  Ensemble classifier \and LIME }

\section{Introduction}
Machine Learning (ML) tasks often involve the training of a model based on past experience and data, which are then used for prediction and classification purposes. The practical applications where such models are used include, e.g.,  loan grants in view of framing laws, detecting terrorism, predicting criminal recidivism, and similar social and economic issues at a global level~\cite{maes2002credit,petercredit,iskandar2017terrorism}. These decisions affect human life and may have undesirable impacts on vulnerable groups in society. The widespread use of ML algorithms has raised multiple concerns regarding user privacy, transparency, fairness, and trustfulness of these models. In order to make Europe 
``fit for the digital age''\footnote{https://www.zdnet.com/article/gdpr-an-executive-guide-to-what-you-need-to-know/}, in 2016 the European Union has enforced the  GDPR Law\footnote{General Data Protection Regulation (GDPR): https://gdpr-info.eu/} across all organizations and firms. The law entitles European citizens the right to have a basic knowledge regarding the inner workings of automated decision models and to question their results. The unfair automated decisions not only violate anti-discrimination laws, but they also undermine public trust in Artificial Intelligence. 
The unwanted bias in the machine learning models can be caused due to the following reasons: 
\begin{itemize}
    \item The \emph{data Collection}~\cite{roh2019survey} may be biased, as certain minority groups of society, or people living in rural areas do not generate enough data. This leads to an unfair model because of unbalanced and biased datasets while training.
    
    \item The \emph{training algorithm} may be subject to bias if one chooses an inappropriate model or training set. Additionally, the model may consider sensitive or discriminatory features while training, which leads to process unfairness.\footnote{Terms unfairness and bias are used interchangeably.}
\end{itemize}
Till now, the notions of fairness have focused on the outcomes of the decision process~\cite{speicher2018unified,zafar2017fairness}, with lesser attention given to the process leading to the outcome~\cite{grgic2018beyond,grgic2016case}. These are inspired by the application of anti-discrimination laws in various countries, which ensures that the people belonging to sensitive groups (e.g. race, color, sex etc.) should be treated fairly. This issue can be addressed through different points of views, which include: \begin{itemize}
    \item \emph{Individual Fairness or Disparate Treatment}~\cite{speicher2018unified} considers individuals who belong to different sensitive groups, yet share similar non-sensitive attributes and require them to have same decision outcomes. For instance, during job applications, applicants having same educational qualifications must not be treated discriminately based on their sex or race.
    \item \emph{Group Fairness or Disparate Impact}~\cite{speicher2018unified} states that people belonging to different sensitive attribute groups should receive beneficial outcomes in similar proportions. In other words, it states that ``Different sensitive groups should be treated equally".
    \item \emph{Disparate Mistreatment or Equal Opportunity}~\cite{zafar2017fairness} proposes different sensitive groups to achieve similar rates of error in decision outcomes.
    \item \emph{Process or Procedural fairness}~\cite{grgic2016case,grgic2018beyond} deals with the process leading to the prediction and keeps track of input features used by the decision model. In other words, the process fairness deals at the algorithmic level and ensures that the algorithm does not use any sensitive features while making a prediction.
\end{itemize}
In this study, we aim to deliver a potential solution to deal with the process fairness in ML Models. The major problem while dealing with process fairness is the opaqueness of ML models. 
Indeed, this black-box nature of ML models, such as in deep neural networks and ensemble architectures such as random forests (RF), makes it difficult to interpret and explain their outputs, and consequently for users and general public to trust their results. There are several proposals of explanatory models to make black-box models more interpretable and transparent.  Due to the complexity of recent black-box models, it is unreasonable to ask for explanations that could represent the model as a whole. This fact, lead to local approaches to derive possible explanations. 

The basic idea is to explain the model locally rather than globally. An ideal model explainer should contain the following desirable properties~\cite{lime}: 
\begin{itemize}
    \item \emph{Model-Interpretability}: The model should provide a qualitative understanding between features and targets. The explanations should be easy to understand.
    \item \emph{Local Fidelity}: It is not possible to find an explanation that justifies the black-box's results on every single instance. But the explainer must at least be locally faithful to the instance being predicted. 
    \item \emph{Model Agnostic}: The explainer should be able to explain all kinds of models.
    \item \emph{Global Perspective}: The explainer should explain a representative set to the user, such that the user has a global understanding of the explainer.
\end{itemize}
Such local explanatory methods include LIME, Anchors, SHAP and DeepSift~\cite{explaining-lime,lime,anchors,shap}. These are based on ``linear explanatory methods'' that gained a lot of attention recently, due to their simplicity and applicability to various supervised ML scenarios.

In this study, we will mainly use LIME 
to derive local explanations of black box classification models. Given a black box model and a target instance, LIME learns a surrogate linear model to approximate the black-box model in a neighbourhood around the target instance. The coefficients of this linear model correspond to the features' contributions to the prediction of the target instance. Thus \emph{LIME outputs top features used by the black box locally and their contributions.} 
In this paper, we propose LIME\textsubscript{Global}, a method to derive global explanations from the locally important features obtained from LIME. 

The LIME\textsubscript{Global} explanations can provide an insight into process fairness. This naturally raises the question of how to guarantee a fairer model given these explanations, while ensuring minimal impact in accuracy~\cite{zafar2015fairness}. This motivated us to seek models $M\textsubscript{final}$ in which $(i)$ their dependence on sensitive features is reduced, as compared to the original model, and $(ii)$ their accuracy is improved (or, at least, maintained).

To achieve both goals, we propose LimeOut\footnote{The name comes from drop-out techniques~\cite{Gal2,Gal1} in neural networks. The github repository of LimeOut can be found here:\\\url{https://github.com/vaishnavi026/LimeOut}}, a framework that relies on {\it feature dropout} to produce a pool of classifiers that are then combined through an ensemble approach. Feature drop out receives a classifier and a feature $a$ as input, and produces a classifier that does not take $a$ into account. Essentially, feature $a$ is removed in both the training and the testing phases.

LimeOut's workflow can be described as follows. Given the classifier provided by the user, LimeOut uses LIME\textsubscript{Global} to assess the fairness of the given classifier by looking into the contribution of each feature to the classifier's outcomes. If the most important features include sensitive ones, the model is unfairly biased. Otherwise, the model is considered as unbiased. In the former case, LimeOut applies dropout of these sensitive features, thus producing a pool of classifiers (as explained earlier).  These are then combined into an ensemble classifier $M_{final}$. Our empirical study was performed on two families of classifiers (logistic regression and random forests) and carried out on real-life datasets (Adult and German Credit Score), and it  shows that both families of models become less dependent on sensitive features (such as sex, race, marital status, foreign worker, etc.) and show improvements or no impact on accuracy.  

The paper is organised as follows. In Section~\ref{sec:related-work} we will discuss some substantial work related to explainability and fairness. We will briefly recall LIME (Local Interpretable Model Agnostic Explanations) in two distinct settings (for textual and tabular data) in Subsection~\ref{sssec:LIME}, and briefly discuss different  fairness issues, some measures proposed in the literature, as well as the main motivation of our work in Subsection~\ref{sssec: Model Fairness}. We will then present our approach (LimeOut) in Section~\ref{sec:methodology}, and  two empirical studies are carried out in Section~\ref{sec:experiments} that indicate the feasibility of LimeOut. Despite the promising results, this preliminary study deserves further investigations, and in Section~\ref{sec:conclusion} we will discuss several potential improvements to be carried out in future work.

\section{Related Work}\label{sec:related-work}
In this section, we briefly recall LIME and and discuss some issues related to model fairness. There has been substantial work done in the field of ``Interpretable Machine Learning" and ``Fairness''. LIME~\cite{lime} and Anchors~\cite{anchors} are prominently being used to obtain the explanations of the black box ML models. These methods provide the top important features that are used by the black box to predict a particular instance. LIME and Anchors do not provide human like explanations (they provide ``feature importance'' or contributions), and they have some limitations~\cite{explaining-lime}. 
In Section~\ref{sec:methodology} we will use LIME to tackle fairness issues based on relative importance of the features. 

\subsection{LIME - Explanatory Method}\label{sssec:LIME}
 LIME (Local Interpretable Model Agostic Explanations) takes the form of surrogate linear model, which is interpretable and mimics locally the behavior of a black box. 
The feature space used by LIME does not need to be the same as the feature space used by a black box. Examples of representations used by LIME include~\cite{lime}: $(i)$ the binary vector representation of textual data that indicates presence/absence of a word, and $(ii)$ the binary vector which represents presence/absence of contiguous patch of similar pixels, in case of images. 

 LIME can be described as follows~\cite{lime}.
Let $f: \mathbb{R}^d \rightarrow \mathbb{R}$ be the function learned by a  classification or regression model over training samples. No further information about this function $f$ is assumed. Now, let $x \in \mathbb{R}^d$ be an instance, and consider its prediction  $f(x)$. LIME aims to explain the prediction $f(x)$ locally.
Note that the feature space of LIME need not be the  same as the input space of $f$. For example, in case of text data interpretable space is used as vectors representing presence/absence of words, whereas the original space might be the word embeddings or word2vec representations. Indeed, LIME uses discretized features of smaller dimension $\hat{d}$ to build the local model, and aims to learn an  explanatory model $g: \mathbb{R}^{\hat{d}} \rightarrow \mathbb{R}$, which approximates $f$ in the neighborhood of $x \in \mathbb{R}^d$.  
To get a local explanation, LIME generates neighbourhood points around an instance $x$ to be explained and assigns a weight vector to these points. The weight is assigned using $\pi_{x}(z)$, which denotes the proximity measure of $z$ w.r.t. $x$. It then learns the weighted linear surrogate model $g$ by solving the following optimisation problem:
\[ g = \mathit{argmin}_{g \in \mathcal{G}}{\;\mathcal{L}(f, g, \pi_{x}(z))} + \Omega(g) \]
where $L(f,g,\pi_{x}(z))$ is a measure of how unfaithful $g$ is in approximating $f$ in the locality defined by $\pi_{x}(z)$, and where $\Omega(g)$ measures the complexity of $g$ (LIME uses the regularization term to measure complexity). In order to ensure both interpretability and local fidelity, LIME  minimizes  $L(f,g,\pi_{x}(z))$ while enforcing $\Omega(g)$ to be small in order to be interpretable by humans. 
\begin{figure}
    \includegraphics[scale = 0.4]{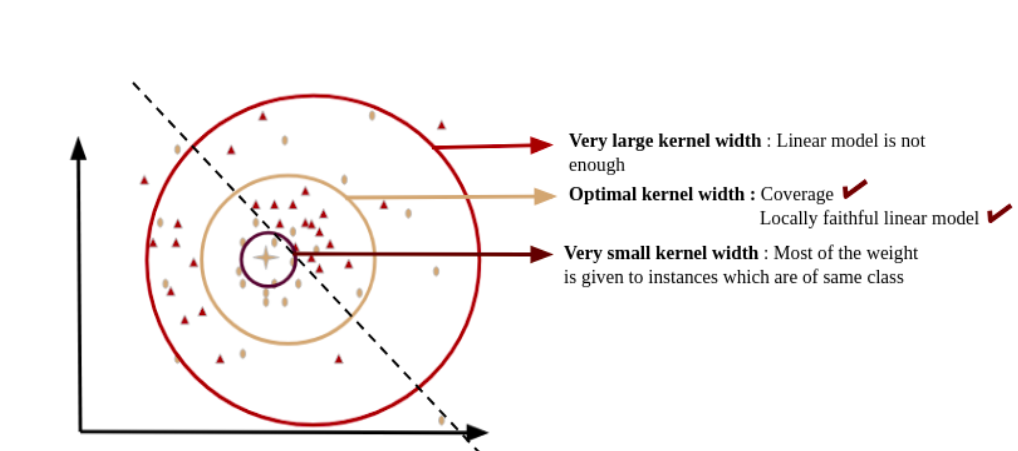}
    \caption{Depicts the $\sigma$'s selection and data distribution, where the red triangles are negative examples, whereas yellow dots constitute positive examples.}
    \label{fig:mesh1}
\end{figure}
The coefficients of $g$ correspond to the contribution of each feature to the prediction $f(x)$ of $x$. LIME uses the following weighting function \begin{equation}\label{eq:weighting}
\pi_{x}(z)=e^{(\frac{−D(x, z)^2}{\sigma^2} )},
\end{equation}
where $D(x, z)$ is the Euclidean distance between $x$ and $z$, and $\sigma$ is the hyper parameter (kernel-width).  The value of $\sigma$ impacts the fidelity of explanation~\cite{laugel2018defining}. For instance, when $\sigma$ is too large, all instances are given equal weight, and it is impossible to derive a linear model which can explain all of them. Similarly if $\sigma$ is too small, only a few points are assigned considerable weight and even a constant model will be able to explain these points, this will result in lower coverage. Thus we need to choose an optimal $\sigma$ to ensure coverage as well as local fidelity (faithfulness). This is illustrated in Figure~\ref{fig:mesh1}: it displays the impact of $\sigma$ on the explanations. The tuned value used by LIME~\cite{lime} for tabular data is $\sigma= 0.75 * n$ for $n$ columns, whereas for textual data it is $\sigma = 25$.

\subsubsection{LIME for textual data~\cite{lime}.}\label{sssec: LimeText}
Consider the text classification problem, in which the goal is to classify an amazon review into positive or negative feedback\footnote{\url{https://www.kaggle.com/bittlingmayer/amazonreviews}}. The model is trained using Naive Bayes Classifier. Let's discuss the procedure to get the LIME explanation:
\begin{enumerate}
 \item\emph{Take any instance $x$ for which you need an explanation}.
    Consider the textual instance \textit{Great easy to set up. Little difficult to navigate and the instructions are non-existent}, and suppose that the Naive Bayes prediction is $P(pos.) = 0.68$ and $P(neg.) = 0.32$. 
 \item\emph{Perturb your dataset and get their black box predictions}.
      For finding the perturbation of this example, LIME randomly removes each word from the original instance (i.e., changes `1' to `0' in the binary representation) one by one, and considers all thus obtained neighborhood points. 
 LIME then gets the black box prediction of these neighbour instances. 
 
 \item\emph{Weight the new samples based on their proximity to the original instance}.
    LIME assigns weights to the neighbourhood instances $z$ based on their proximity to the original instance $x$ using \ref{eq:weighting}.
 \item\emph{Fit a weighted, interpretable (surrogate) model on the dataset with the variations.}
    LIME trains a linear weighted model that fits the original and the obtained neighbourhood instances.
 \item\emph{Get the explanations by interpreting the local model.}
    The output of LIME is the list of explanations, reflecting the contribution of each feature to the prediction of the sample.
    The resulting explanation is illustrated in Figure~\ref{fig:mesh3}
    \begin{figure}

        \includegraphics[scale = 0.4]{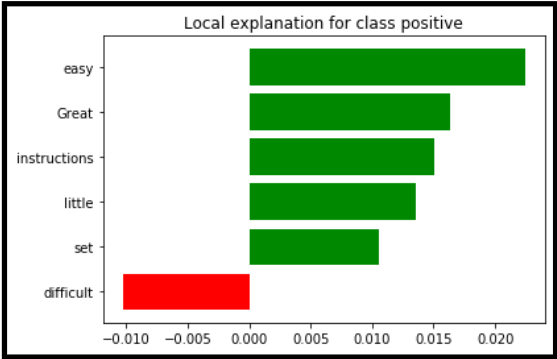}
        \caption{The explanation for the classification of
        {\it Great easy to set up. Little difficult to navigate and the instructions are non-existent}, which indicates the contribution of each word (in red is the contribution to the negative feedback class, and in green to the positive feedback class).}
        \label{fig:mesh3}
\end{figure}
\end{enumerate}

\subsubsection{LIME for tabular data~\cite{explaining-lime}.} \label{sssec: LIMETabular}
The workflow of LIME on tabular data is similar to that on textual data. However, unlike LIME for textual data, it needs a training set (user defined) to generate neighbourhood points. The following statistics are computed for each feature depending on their type: $(i)$ for categorical features it computes the frequency of each value, $(ii)$ for numerical features, it computes the mean and the standard deviation, which are then discretized into quartiles.

\begin{figure}
        \includegraphics[scale = 0.4]{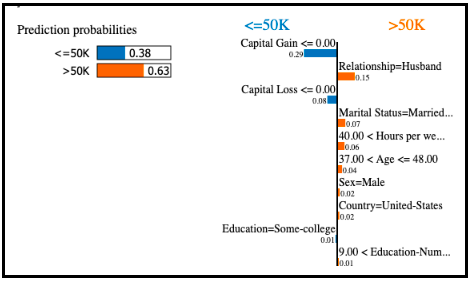}
        \caption{Local explanation in case of Adult dataset. The orange bar represents the contribution of feature, to predict salary $\geq$ 50k dollars and blue bar is for the features which contributes to the negative class (salary $<$ 50k dollars)} 
        \label{fig:mesh4}

\end{figure}

Suppose that $f$ is the black-box function, and that we want to explain the prediction $f(x)$ of $x$ = $(x_1, x_2, \ldots, x_i., x_n)$, where each $x_i$ may be a categorical or a numerical value. Each categorical value is mapped to an integer using LabelEncoder\footnote{LableEncoder Class is given in the sklearn preprocessing library\\\url{https://scikit-learn.org/stable/modules/generated/sklearn.preprocessing.LabelEncoder.html}}. 
Note that the values of each feature in the training set is divided into $p$ quantiles. These quantile intervals are used for discretizing the original instance. If $x_i$ lies between quantile $q_j$ and $q_{j+1}$, it gets the value $j$. This is done for all the features to get the quantile boxes for all $x_i,  i \in \{1,\ldots,n\}$. 
\\
To get the perturbation $\hat{y}$ in the neighbourhood of $\hat{x}$, LIME samples discrete values from $\{1,\ldots,p\}$, $n$ times.  To get the continuous representation $y$ of $\hat{y}$, LIME Tabular uses a normal distribution and the quantile values. The neighbourhood instance $\hat{y}$ is represented as binary tuple with the $i$-th component equal to 1 if $\hat{x_i} = \hat{y_i}$, and 0 if $\hat{x_i} \neq \hat{y_i}$.
In this way LIME Tabular generates all the neighbourhood points. The following steps are similar to LIME for textual data. These points are assigned weights using the exponential kernel~(\ref{eq:weighting}), and a weighted linear function is learned over the neighbourhood permutations.
To illustrate, consider an example of the Adult dataset (see Subsection~\ref{sssec:Adult}). The task is to predict if a salary of a person is $\geq$ 50k dollars. We have trained the model using Random Forest Classifier. An example of local explanation is given in Figure~\ref{fig:mesh4}.

\subsection{Model Fairness}\label{sssec: Model Fairness}
Several notions of model fairness have been proposed ~\cite{dressel2018accuracy,speicher2018unified,zafar2017fairness,grgic2018beyond,grgic2016case} based on decision outcomes as well as on process fairness. Individual fairness~\cite{chouldechova2017fair} (or disparate treatment, or predictive parity) imposes that the instances/individuals belonging to different sensitive groups, but similar non-sensitive attributes must receive equal decision outcomes. The notion of group fairness(or disparate impact or statistical parity
~\cite{dwork2012fairness}) is rooted in the desire for different sensitive demographic groups to experience similar rates of errors in decision outcomes. COMPAS\footnote{ \url{https://en.wikipedia.org/wiki/COMPAS\_(software)}} is a recidivism detection tool, where the goal is to predict whether a criminal would re-offend his crime based on a long questionnaire. The then popular algorithm was designed by the commercial company, Northpointe (now Equivant). A study by ProPublica\footnote{\url{https://www.propublica.org/article/machine-bias-risk-assessments-in-criminal-sentencing}} showed that COMPAS has a strong ethnic bias. Among non-reoffenders, COMPAS is almost twice more likely to signal black people as high risk. Furthermore in COMPAS, white reoffenders are predicted as low risk much often than black offenders. In other words, this indicates that COMPAS has considerable high false positive and lower true negative rates for black defendants when compared to white defendants. COMPAS is used across US by judges and parole officers to decide whether to grant or deny probation to offenders; hence, it is very important to understand how this model reaches its conclusion and ensure it is fair. 
If we focus on the decision outcomes, the fair algorithm in case of COMPAS (if we consider only Race as sensitive feature) should be such that: $(i)$ black and whites with the same features get the same output (no disparate treatment and thus non-discriminatory), and $(ii)$
the proportion of individuals classified as high-risk should be same across both the groups (statistical parity).

We can deal with this bias during training (see \cite{zafar2017fairness}) by:
$(i)$ excluding all features that may cause the model to create bias, e.g race, gender etc., or $(ii)$ including discrimination measures as learning constraints, i.e., the model should be trained to minimize  $P(y_{{pred}}  \neq y_{{true}})$  such that 
$$P(y_{{pred}}  \neq y_{{true}} | race = Black)  =   P(y_{{pred}}  \neq y_{{true}} | race = White),$$
where $y_{{pred}}$ is the risk predicted by trained ML model (e.g., COMPAS) and $y_{{true}}$ is the true risk value.
This constraint is motivated by the fact that `race' is a sensitive feature. 
Such constraints are applied to different sensitive attributes separately (e.g. sex, race, nationality etc.), it might lead to unfairness for the groups which lie at the intersection of multiple kinds of discrimination (e.g. black women), also known as \emph{fairness gerrymandering}~\cite{kearns2017preventing}. To avoid this, \cite{zhang2016identifying} proposed constraints for multiple combinations of sensitive features. However, constraints for multiple combinations of sensitive attributes render model training highly complex and may lead to overfitting.

Earlier studies in fair ML \cite{zemel2013learning,zafar2015fairness} consider individual and group fairness as conflicting measures, and some studies tried to find an optimal trade-off between them. In~\cite{binns2020apparent} the author argue that, although apparently conflicting, they correspond to the same underlying moral concept. In fact, the author provides a broader perspective and advocates an individual treatment and assessment on a case-by-case basis.
In~\cite{grgic2016case,grgic2018beyond} the author provides another noteworthy perspective to measure fairness, namely, \emph{process fairness}. Rather than focusing on the outcome, it deals with the process leading to the outcome. In~\cite{grgic2018beyond} the author provides a key insight to rely on \emph{human's moral judgement or intuition} about the fairness of using an input feature in algorithmic decision making. He also assesses the impact of removing certain input features on the accuracy of the classifier, and designs an optimal trade-off between accuracy and the process fairness for the classifier. However, humans may have different perspectives on  whether it is fair to use a input feature in decision making process. In~\cite{grgic2018human} the authors propose a framework to understand why people perceive certain features as fair or unfair. They introduce seven factors on which a user evaluates a feature in terms of reliability, relevance, privacy, volitionality, causes outcome, causes vicious cycle, causes disparity in outcomes, caused by sensitive group membership. 

We are inspired by the idea of using a combination of classifiers instead of a single one. For instance, in~\cite{grgic2016case} the authors explore the benefits of replacing a single classifier with a diverse ensemble of random classifies, regarding the accuracy as well as individual and group fairness. 
In this paper, we further explore this idea 
and propose a method, that we call \emph{LimeOut}, to ensure process fairness while improving (or, at least, maintaining) the model's accuracy.

\section{Our Methodology}\label{sec:methodology}
In this section, we describe in detail the framework of LimeOut that consists of two main components: LIME\textsubscript{Global} and ENSEMBLE\textsubscript{Out}. It receives as input both a classifier\footnote{Here we focus on binary classifiers that output the probability for each class label.} and a dataset. The first component then checks whether the classifier is biased on the dataset in the sense that the predictions depend on sensitive features. To do this, we make use of LIME\textsubscript{Global}~\cite{lime} (see Subsection~\ref{sssec:Limeglobal}). This will output the most important features (globally).  If  sensitive features are among the most important, then the classifier is considered unfair and the second component of LimeOut is employed. Otherwise, the classifier is considered fair and no action is taken. The second component is the core of LimeOut (see Subsection~\ref{sssec: Limeout}). Given the most important features, ENSEMBLE\textsubscript{Out} produces a pool of classifiers using feature-drop. Each of these classifiers does not depend on the corresponding sensitive features. It then constructs an ensemble using this pool of classifiers. Following a human and context-centered approach, the choice of sensitive features is left to the user within the given context. This framework will be illustrated in Section~\ref{sec:experiments}.

\subsection{LIME\textsubscript{Global}}\label{sssec:Limeglobal}
LIME is prevalent to get local explanations for the instances. These explanations can be combined to provide insights into the global process of the classifier~\cite{lime,van2019global}. First, LIME\textsubscript{Global} chooses instances using \emph{submodular pick} method~\cite{lime}. The choice of instances can impact the reliability of the global explanation. The method submodular pick provides a set of instances for which explanations are diverse and non-redundant. To obtain a global insight into the classifier's inner process, we use the instances obtained from submodular pick\footnote{In \cite{lime} the authors argue that the submodular pick is a better method than random pick. We still experimented random pick on the datasets of Section~ \ref{sec:experiments}, but the relative importance of features remained similar.}. LIME\textsubscript{Global} obtains the local explanations (important features and their contributions) for all these instances. 
This results in a list of top important features used by the model globally.

\subsection{ENSEMBLE\textsubscript{Out}}\label{sssec: Limeout}
LimeOut uses the globally important features obtained by LIME\textsubscript{Global} to assess process fairness of any given ML model.
In this way, we can check whether the model's predictions depend on sensitive features and measure its dependence. If sensitive features are ranked within the top 10\footnote{In this study we focused on the top 10 features. However this parameter can be set by the user and changed according to his use case.} globally important features, then it is deemed unfair or biased.
If the model is deemed unfair, then one easy solution would be  to remove all the sensitive features from the dataset before training. 
However, these sensitive features may be highly correlated to non-sensitive features, thus keeping the undesired bias. To mitigate this drawback, LimeOut also removes all such correlated features. 

Now this could entail a decrease in performance since,  after removing all the sensitive features, the model could become less accurate due to the lack of training data. To overcome this limitation, LimeOut constructs a pool of classifiers each of which corresponding to the removal of a subset of sensitive features. To avoid the exponential number of such classifiers, in this paper we only consider those obtained by removing either \emph{one} or \emph{all} sensitive features. 
LimeOut constructs an ensemble classifier $M_{final}$ through a linear combination of the pool's classifiers. 


More precisely, given an input $(M,D)$, where $M$ is a classifier and $D$ is the dataset. Suppose that the globally important features given by LIME\textsubscript{Global} are $a_1, a_2$,\ldots,$a_n$, in which $a_{j_1}, a_{j_2},\ldots, a_{j_i}$ are sensitive. LimeOut thus trains $i+1$ classifiers: $M_k$ after removing $a_{j_k}$ from the dataset, for $k=1, \ldots, i$, and $M_{i+1}$ after removing all sensitive features $a_{j_1}, a_{j_2},\ldots, a_{j_i}$.
In this preliminary implementation of LimeOut, the ensemble classifier $M\textsubscript{final}$ is defined as the ``average'' of these $i+1$ classifiers. More precisely, for an instance $x$ and a class $C$,
\[
{P}_{M_{final}}(x\in C) =  \frac{\sum_{k=1}^{k=i+1} {P}_{M_k}(x\in C)}{i+1}.
\]
As we will see empirically in Section~\ref{sec:experiments} over different datasets and classifiers, the dependence of $M_{final}$ on sensitive features decreases, whereas its accuracy is maintained and, in some cases, it even improves. 

\section{Experiments}\label{sec:experiments}
To validate our approach, we applied LimeOut on two different families of classifiers (Logistic regression and Random Forests) over different datasets.  In each case, the ensemble classifier obtained by LimeOut is fairer than the original classifiers. The datasets we use, Adult and German credit score, are known to be biased. These experiments illustrate different possible scenarios, namely, the case of unfair process (see Subsection~\ref{sssec:Adult}) and of a fair process (see Subsection~\ref{sssec: German} for Random Forests).

\subsection{Adult Dataset}\label{sssec:Adult}

This dataset comes from the UCI repository of machine learning databases\footnote{Adult Dataset: \url{http://archive.ics.uci.edu/ml/datasets/Adult}}. The task is to predict if an individual's annual income exceeds 50,000 dollars based on census data. An individual’s annual income is the result of various features such as ``Age'', ``Workclass'', ``fnlwgt'', ``Education'', ``Education\-Num'', ``Marital Status'', ``Occupation'', ``Relationship'', ``Race'', ``Sex'', ``Capital Gain'', ``Capital Loss'', ``Hours per week'' and ``Country''. 
Intuitively, the income of a person should get influenced by the individual’s education level, age, occupation, number of hours he works, company etc. But it would be unfair if our model considers race, sex or the marital status of the individual while making any prediction. 

This  dataset has 14 features out of which 6 are continuous and 8 are nominal, and it comprises 45,255 instances. We partitioned the dataset randomly into 80\% for training and 20\% for testing.
However, the class distribution of Adult dataset is extremely unbalanced and majority of the dataset consists of individuals with annual income $<50,000$ dollars. To balance this, we used  Synthetic Minority Oversampling Technique (SMOTE\footnote{\url{https://imbalanced-learn.readthedocs.io/en/stable/generated/imblearn.over\_sampli\\ng.SMOTE.html}}) over training data. SMOTE generates new samples from the minority class and includes them in the training set, resulting to a balanced training dataset. We then perform training on the augmented (balanced) dataset using: Logistic Regression and Random Forest. 

\subsubsection{Logistic Regression:} 
We trained a logistic regression model over the obtained training set. In  binary classification problems, logistic regression often uses a default threshold value of 0.5, i.e. if predicted value $\geq$ 0.5, then the predicted class will be positive, and negative, otherwise. However, this threshold may lead to poor results, especially, in the case of unbalanced datasets. We used threshold tuning\footnote{\url{https://machinelearningmastery.com/threshold-moving-for-imbalanced-classification/}} in order to improve the performance of our classifier. The threshold is chosen to be optimal for Precision Recall Curve and the ROC Curve (to ensure maximum F1-score). 
The classifier $M$ obtained after threshold tuning had an accuracy of 82.65\%. To assess the process fairness of $M$, we used LIME\textsubscript{Global} to get the 10 most important features used by $M$.


\begin{table}[t]
   \begingroup
        \setlength{\tabcolsep}{5pt} 
        \renewcommand{\arraystretch}{1.5} 
        \begin{minipage}{.5\linewidth}
            \centering
            \begin{tabular}{|c|c|}
                \hline
                \textbf{Features } & \textbf{ Contribution }\\\hline
                Capital Gain & -23.792107 \\\hline
                Capital Loss & -6.469338 \\\hline
                Hours Per Week & -2.496092\\\hline
                \textbf{Marital Status} & 2.116016\\\hline
                \textbf{Race} & 1.927533 \\\hline
                \textbf{Sex} & 1.804058 \\\hline
                Education-Num & -1.573597 \\\hline
                Age & 0.698024\\\hline
                Education & 0.667795\\\hline
                Relationship & 0.235550\\\hline
                
            \end{tabular}
            \label{table:t1}
        \end{minipage}%
        \begin{minipage}{.5\linewidth}
          \centering
            \begin{tabular}{|c|c|}
                \hline
                \textbf{Features } & \textbf{ Contribution }\\\hline
                Capital Gain & -23.543842 \\\hline
                Capital Loss & -5.767617 \\\hline
                Education-Num & -1.673827 \\\hline
                Hours Per Week & -1.541263\\\hline
                Country  & 0.80206 1\\\hline
                Education & 0.547427\\\hline
                \textbf{Sex} & 0.477145 \\\hline
                Workclass & 0.426351\\\hline
                Age & -0.242858\\\hline
                Relationship & 0.065351\\\hline
             
            \end{tabular}
        \end{minipage} 

         \caption{Top 10 important features used by $M$ (left) and $M_{final}$ (right). }
        \label{table:g1}
    \endgroup 
\end{table}
From Table~\ref{table:g1}, it is evident that Race, sex and marital status are among the top 10 features used by model $M$ with contributions 1.93, 1.80 and 2.11 respectively.
We know that it's unfair to use these features while predicting someone's income. And as these are among the top 10 features, we can deem the model to be unfair.
Now we train four models by dropping out sensitive features : Race, Sex and Marital status. Note that all the classifiers are trained using Logistic Regression with threshold tuning. Through feature dropout, we thus obtain 4 classifiers:
 $M1$ trained without ``Sex'', 
 $M2$ trained without ``Race'',
 $M3$ trained without ``Marital Status'', and
 $M4$ trained without the 3 (Accuracy = 81.97\%).

We can infer that $M4$ is fairer because it has not used any sensitive feature while training. But the accuracy is reduced from 82.65\% to 81.9\%. 
The ensemble $M\textsubscript{final}$ of models $M1$, $M2$, $M3$ and $M4$
achieved an accuracy of 84.18\%. The statistical test\footnote{We performed the t-test.} showed that this improved accuracy is significant. The global impact of the sensitive features is also reduced (see the explanations in Table \ref{table:g1}). 

\subsubsection{Random Forest:}
We also used Random Forest and checked its fairness. This model $M_{RF}$  has accuracy = 83.49\%. The global explanations for $M_{RF}$ LimeOut's ensemble model $(M_{RF})_{final}$ are given in Table~\ref{table:g2}. 
\begin{table}[t]
   \begingroup
        \setlength{\tabcolsep}{5pt} 
        \renewcommand{\arraystretch}{1.5} 
        \begin{minipage}{.5\linewidth}
            \centering
            \begin{tabular}{|c|c|}
                \hline
                \textbf{Features } & \textbf{ Contribution }\\\hline
                Capital Gain & -10.218573 \\\hline
                Capital Loss & -3.109039 \\\hline
                Hours Per Week & -1.332370\\\hline
                \textbf{Sex} & 1.244931 \\\hline  
                \textbf{Marital Status} & 0.744446\\\hline
                \textbf{Race} & 0.456074 \\\hline
                Occupation & -0.256388\\\hline
                Age & -0.249529 \\\hline
                Country & 0.249083\\\hline
                Relationship & 0.215706\\\hline
            \end{tabular}
        \end{minipage}%
        \begin{minipage}{.5\linewidth}
          \centering
            \begin{tabular}{|c|c|}
                \hline
                \textbf{Features } & \textbf{ Contribution }\\\hline
                Capital Gain & -10.304901 \\\hline
                Capital Loss & -3.436587\\\hline
                Hours Per Week & -1.362630\\\hline
                Education-Num & 0.574524\\\hline
                Relationship & 0.413276\\\hline
                \textbf{Sex} & 0.306334\\\hline
                \textbf{Marital Status} & 0.243644\\\hline
                Workclass & 0.137123 \\\hline
                Country & 0.091939\\\hline
                Occupation & 0.078968\\\hline

            \end{tabular}
        \end{minipage} 
        
         \caption{Top 10 important features used by $M_{RF}$ (left) and $(M_{RF})_{final}$ (right). }
        \label{table:g2}

    \endgroup 
\end{table}
From the Table \ref{table:g2} we see that the impact of sensitive features  decreased for $(M_{RF})_{final}$, and that its accuracy increased to 83.86\%. While when we removed all three sensitive features Race, Sex and Marital status, the accuracy was 81.6\%. Again we observe a significant improvement in the accuracy of the LimeOut's ensemble classifier, while ensuring a fairer model.

\subsection{German Credit Score Dataset}\label{sssec: German}

The data was initially prepared by Prof. Hoffman and is available publicly as `german.data' on UCI Machine Learning Repository \footnote{\url{https://archive.ics.uci.edu/ml/datasets/statlog+(german+credit+data)}}. If a bank receives a loan application based on the applicant's profile it can decide whether it can approve the loan. Two types of risk are associated with the bank's decision: $(i)$ if an applicant is at \emph{good credit risk}, he is likely to pay back his loan, and $(ii)$ if an applicant is at \emph{bad credit risk}, he is unlikely to pay back.

The dataset set has information about 1000 individuals on the basis of which they have been classified as good or bad risk. The goal is to use applicant's demographic and socio-economic profiles to assess the risk of lending loan to the customer. The dataset consists of 20 features and a classification label (1: Good Risk, 2: Bad Risk). 
We split the dataset into 80\% training set and 20\% testing. As the dataset is highly imbalanced, we used SMOTE Oversampling to generate the samples synthetically.

\begin{table}[t]
  \begingroup
        \setlength{\tabcolsep}{5pt} 
        \renewcommand{\arraystretch}{1.5} 
        \begin{minipage}{.5\linewidth}
            \centering
            \begin{tabular}{|c|c|}
                \hline
                \textbf{Features } & \textbf{ Contribution }\\\hline
                peopleliable & -6.410473 \\\hline
                \textbf{foreignworker} & 5.398807\\\hline
                otherinstallmentplans & -1.769830 \\\hline
                savings & 1.769533 \\\hline
                \textbf{telephone} & 1.349587 \\\hline
                \textbf{statussex} & -1.263993 \\\hline
                creditamount & 0.899089 \\\hline
                existingchecking & 0.798037\\\hline
                duration & 0.691543 \\\hline
                employmentsince & -0.619419\\\hline
                 
            \end{tabular}
            \vspace{2mm}
            
        \end{minipage}%
        \begin{minipage}{.5\linewidth}
          \centering
            \begin{tabular}{|c|c|}
                \hline
                \textbf{Features } & \textbf{ Contribution }\\\hline
                peopleliable & -5.210576 \\\hline
                \textbf{foreignworker} & 2.586505 \\\hline
                otherinstallmentplans & -1.858603 \\\hline
                credithistory & 1.418544 \\\hline
                installmentrate & 1.185539 \\\hline
                savings & 1.087709 \\\hline
                purpose & 0.570004 \\\hline
                duration & -0.427534 \\\hline
                employmentsince & -0.385297\\\hline
                creditamount & 0.354635 \\\hline
                
            \end{tabular}
            \vspace{2mm}
        \end{minipage} 
         \caption{Top 10 important features used by $M$ and $M\textsubscript{final}$ respectively \emph{(Logistic Regression)}. These explanations are obtained using LIME\textsubscript{global}.}
        \label{table:g3}
    \endgroup 
\end{table}

\subsubsection{Logistic Regression:}
For training we used Logistic Regression along with threshold tuning. The obtained accuracy of $M$ was 74.67\% with the explanations from LIME\textsubscript{Global} given in Table~\ref{table:g3}. Here, we see the sensitive features ``statussex'' (sex of the customer), 
``telephone\footnote{It depicts if a person has registered his telephone number. Due to privacy reasons, someone may not share his number. Thus this features should not be considered important}'' and ``foreign worker'' appear in the top 10, thus showing  that $M$ is process unfair. Hence, LimeOut trains $M1$, $M2$ and $M3$ by removing  each one of them, and $M4$ after removing all 3. Despite being fairer, $M4$ suffered a drastic accuracy decrease to 69\%.

LimeOut then trained the ensemble $M\textsubscript{final}$ and output the explanations given in Table~\ref{table:g3}. Again, the impact of sensitive features decreased in case of $M\textsubscript{final}$. In addition, the accuracy of M\textsubscript{final} is 74.67\%, same as $M$. Again, a fairer classifier without compromising accuracy.

\subsubsection{Random Forest} 
We trained the model using Random Forest, and the accuracy was found to be 59\%. In this case, LIME\textsubscript{Global}
showed a single sensitive feature in the top 10 and no action was taken\footnote{Interestingly, there is an accuracy increase when that variable is dropped. However, the current implementation of LimeOut does not take action in these cases.}.
We will further discuss this case in the next section.

\section{Conclusion and Future Work}\label{sec:conclusion}
We demonstrated the idea of using LIME to determine model fairness, 
and integrated it in LimeOut that receives as input a pair $(M,D)$ of a classifier $M$ and a dataset $D$, and outputs a classifier $M_{final}$ less dependent on sensitive features without compromising accuracy.

This preliminary study shows the feasibility and the flexibility of the simple idea of feature dropout followed by an ensemble approach. This opens into several potential improvements and further investigations. First, we only experimented LimeOut on two classes of classifiers, but LimeOut can be easily adapted to different ML models and data types, as well as different explanatory models. An improvised approach to get the global explanation like \cite{van2019global} can be used, and this should be thoroughly explored. 

Also, the workflow can be further improved, e.g., the classifier ensembles could take into account classifier weighting and other classifiers resulting from the removal of different subsets of sensitive features (here we only considered the removal of one or all features).
In this study, we took a human and context-centered approach that requires domain expertise (for identifying sensitive features in a given use-case). However, there is room for automating this task, possibly through a metric or utility-based approach to assess sensitivity  that takes into account domain knowledge. 

We also identified some limitations as that illustrated in the last scenario. Indeed, despite providing insights on process fairness, LimeOut seems of little use when only one sensitive feature is detected in the top $k$ important features. In this case, an alternative method should be employed, for instance, to consider the model obtained by removing this feature. These are some of the issues to be tackled in future work.  

\bibliographystyle{unsrt}  

\bibliography{references}

\end{document}